\pdfoutput = 1
\PassOptionsToPackage{hidelinks}{hyperref} 
\PassOptionsToPackage{square}{natbib} 
\documentclass[letterpaper]{article}

\usepackage{natbib,alifeconf} 
\usepackage{booktabs}
\usepackage{config} 
\usepackage{url,hyperref,cleveref} 

\title{On Creativity and Open-Endedness}

\author{
L. B. Soros$^{1}$, 
Alyssa M. Adams$^{3}$, 
Stefano Kalonaris$^{2}$, 
Olaf Witkowski$^{3-6}$, \and 
Christian Guckelsberger$^{7,8}$\\
\mbox{}\\
$^1$Department of Computer Science, Barnard College, USA
$^2$Riken AIP, Japan
$^3$Cross Labs, Cross Compass Ltd., Japan\\
$^4$Graduate School of Arts and Sciences, University of Tokyo, Japan\\
$^5$Earth-Life Science Institute, Tokyo Institute of Technology, Japan
$^6$Institute for Advanced Study, USA\\
$^7$Department of Computer Science, Aalto University, Finland\\
$^8$School of Electronic Engineering and Computer Science, Queen Mary University of London, UK\\
christian.guckelsberger@aalto.fi
} 
\begin{document}

\maketitle

\begin{abstract}
 \ac{ALife} as an interdisciplinary field draws inspiration and influence from a variety of perspectives. Scientific progress crucially depends, then, on concerted efforts to invite cross-disciplinary dialogue. The goal of this paper is to revitalize discussions of potential connections between the fields of \ac{CC} and \ac{ALife}, focusing specifically on the concept of \ac{OE}; the primary goal of \ac{CC} is to endow artificial systems with creativity, and \ac{ALife} has dedicated much research effort into studying and synthesizing \ac{OE} and artificial innovation. However, despite the close proximity of these concepts, their use so far remains confined to their respective communities, and their relationship is largely unclear. We provide historical context for research in both domains, and review the limited work connecting research on creativity and \ac{OE} explicitly. We then highlight specific questions to be investigated in future work, with the eventual goals of (i) decreasing conceptual ambiguity by highlighting similarities and differences between the concepts of \ac{OE} and creativity, (ii) identifying synergy effects of a research agenda that encompasses both concepts, and (iii) establishing a dialogue between \ac{ALife} and \ac{CC} research. 
 
\end{abstract}

\section{Introduction}
\label{sec:intro}

This paper seeks connections between two fields, communities, and research endeavours: \acf{ALife} and \acf{CC}. \ac{ALife} researchers study \enquote{life and life-like processes through simulation and synthesis}~\citep{bedau2007artificial} in software, hardware, and wetware, whereas \ac{CC} researchers are concerned with the theory and engineering of computational systems capable of exhibiting creativity, including systems that co-create with others \citep[e.g.][]{colton2012computational}. With regards to \ac{ALife}, we focus on what has been considered a grand challenge \citep{bedau2000open} and major goal \citep{stepney2021modelling}  of the field: the definition, categorization, and measurement of \acf{OE} as a characteristic of life and natural evolution. 

The engineering of both \ac{CC} and \ac{OE} have equally been advocated as the \enquote{final frontier} \citep{colton2012computational}, and the \enquote{last grand challenge} \citep{stanley2017open} of \ac{AI} research, respectively. In AI research more generally, \ac{OE} has been claimed to be essential for any artificial superhuman intelligence 
\citep{hughes2024openendedness}. These claims, together with an immense heritage and growing body of related work, highlight the magnitude of, and desire toward accomplishing, these two research challenges. At the basis of this paper lies the observation that neither challenge is close to being fully solved, but, at the same time, both challenges appear related. We consequently ask: Are creativity and \ac{OE} entirely different concepts, or do their synthesis and measurement represent closely related research endeavors? And if so, how are they connected? 

There is much reason to believe in the latter. Creativity, more broadly, is understood as a hallmark feature of human \citep{still2016history} and (with nuances) animal cognition \citep{kaufman2015animal}, while \ac{OE} is considered foundational for the evolution and development of advanced lifeforms \citep{bedau2000open}. In creativity studies \citep{williams_2016_mapping_creativity_research}, \ac{OE} is often used as a property of tasks and tests to assess human creativity \citep{epstein2011exercises} and in \ac{CC}, it is used as challenge to motivate the advancement of algorithms \citep[e.g.][]{guckelsberger2016supportive}. Both creativity and \ac{OE} have been proposed together as features of (especially evolutionary) algorithms across both disciplines \citep[e.g.][]{stanley2015greatness}. Moreover, creativity is very often mentioned together with, or amalgamated in, definitions of \ac{OE} \citep{bedau1998four, packard2019openended, hughes2024openendedness}, albeit without further differentiation.

We argue that these observations signal opportunity: there is potential for the \ac{CC} and \ac{ALife} communities to learn about each other's research heritage and challenges, and, especially in close interaction, use it to advance research on either front. Making use of this opportunity is important, since research involving both creativity and \ac{OE} has unresolved ambiguities in the involved concepts, including their definitions and measurements. In a way, this also marks an (unfortunate) commonality between the two areas. 

For this paper, we brought together a team of \ac{CC} and \ac{ALife} researchers to make three distinct contributions: we (1) summarise previously identified connections between \ac{CC} and \ac{OE} from isolated, existing work; we (2) uncover new connections that have the potential to promote further work based on our expert discussions; and we (3) identify synergy effects of a research agenda that encompasses both with the goal to spark discussion and interdisciplinary collaboration between  \ac{CC}, \ac{ALife}, and other communities. 

While our goal is ultimately to deepen our understanding of each concept and the requirements to reproduce them in artificial systems, we do not pretend to resolve these connections conclusively given limited space, time, and expertise. Instead, we ask provoking questions on the nature and constituents of these connections as \textit{starting points} for additional deep, dedicated interdisciplinary investigations. 

By doing so, we revitalize a prior discussion by \citet{sorosevolution2016}, asking ``Are there nontrivial differences between creativity and OEE, or are they fundamentally the same concept?'' 
Our treatment is more extensive and general, as it also considers \ac{OE} beyond \ac{OEE}. 
Existing work in \ac{CC} focuses on \ac{OE} as a property of tasks / exercises / problems \citep[e.g.][]{epstein2011exercises} or environments \citep[e.g.][]{guckelsberger2016supportive}, but not as a characteristic of the process or the resulting product which \ac{ALife} seeks to reproduce.~Vice versa, \ac{ALife} researchers have drawn connections from \ac{OE} to creativity, but articulated only as sidelines. Instances of work spanning both fields and research agendas \citep{boden2009evolution,sorosevolution2016} are promising, but few and isolated. 

In a more general sense, this paper continues a long-lasting tradition of knowledge exchange between the fields of \ac{ALife} and \ac{CC}. While it was written primarily with the goals and understanding of \ac{CC} and \ac{ALife} researchers in mind, creativity and \ac{OE} are becoming dramatically more relevant beyond the boundaries of these communities, especially for AI more broadly \citep[e.g.][]{hughes2024openendedness}. This paper is written for a wider audience, and we hope that our insights can inspire and advance research in \ac{AI} and beyond.

\section{Background}
\label{sec:background}

The concepts of \ac{OE} and creativity (more broadly) are used in a multitude of fields and contexts, such as educational research \& psychology, games \& play, software engineering, and design, just to name a few. Here, we focus on their use specifically within the fields of \acf{ALife} and \acf{CC}, separate from adjacent fields such as machine learning and AI research more generally. At the same time, we acknowledge that the boundaries between fields are fluid and evolving\footnote{E.g.~the workshops on \enquote{Agent Learning in Open-Endedness} (ALOE) and \enquote{Intrinsically Motivated Open-ended Learning} (IMOL) at NeurIPS also feature \ac{CC} and \ac{ALife} research.}. As a first notable difference between both concepts, they play a very different role in the definition of the respective discipline's identity. We next introduce each discipline in more detail, together with the core concepts of creativity, \ac{OE} and \ac{OEE}.

\subsubsection{(Computational) Creativity}

\ac{CC} has been defined as, amongst others, \enquote{the art, science, philosophy, and engineering of computational systems which, by taking on particular responsibilities, exhibit behaviors that unbiased observers would deem to be creative}~\citep{colton2012computational}. This places creativity right at the core of the discipline's identity. Researchers follow two broad goals: \enquote{understanding human creativity} and \enquote{producing machine creativity} \citep{boden2004creative, perez2018continuum}. The latter can be differentiated into co-creative systems \citep{kantosalo2016modes}, taking on some creative responsibility \citep{colton2012computational} in interaction with others, and fully autonomous creative systems \citep{colton2008creativity} -- marking two extremes on an interaction continuum. 
A deeper introduction of \ac{CC} as a field is beyond the scope of this paper, and we refer the reader to excellent overview articles of its early history and academic predecessors \citep{boden2015history} as well as more recent depiction \citep{veale2019systematizing}. Crucially, the goals introduced above are increasingly also pursued in other fields and under different labels such as Creative AI and Generative AI, amongst others \citep{vear2022creativeAI}. 

\ac{CC}'s conceptualization of creativity is mostly adopted from (psychological) creativity studies, the long-standing joint effort of psychology, sociology, philosophy, and other non-computational disciplines to investigate creativity \citep{williams_2016_mapping_creativity_research}. \ac{CC}'s ontological contribution then consists of re-evaluating and extending the scope and attribution of creativity in the context of (typically computational) artificial systems. Similarly to creativity studies, \ac{CC} follows an anthropocentric agenda, and creativity is almost always connected to features of human creativity. 

Creativity is a very old concept, and our understanding today mixes two historical traditions \citep{still2016history}. In the older, Pagan tradition, creativity concerns the unfolding and dissolution of natural processes, such as the birth of a child or the growth of a plant. Here, \emph{create} translates to \enquote{having an impact through natural forces} [ibid] and is different from \emph{facere}, which corresponds to \enquote{make out of available materials} [ibid]. The younger, Christian tradition used \emph{creare} as \enquote{bringing about by making} [ibid], thus mixing the two prior Pagan notions. It is typically considered an individual power, denoting \enquote{creation out of nothing} [ibid], and has mostly been attributed to the genius of artists who bring forth novel ideas, rather than exercising skill alone. 

Following a surge in creativity-related research starting in the 1960s, definitions of creativity proliferated. Already in \citeyear{taylor1988various}, 50 different definitions had been proposed over the previous five decades. This development prompted researchers to lament that either creativity has \enquote{almost ceased to mean anything} \citep{batey2006creativity} or is an essentially contested concept \citep{jordanous2016modelling}. 

\citet{runco2012standard} have helped mitigate this situation by abstracting common underlying factors in existing studies and definitions, arriving at a two-component \enquote{standard definition} of creativity which requires a product or process to be both \emph{novel} and \emph{valuable} to be deemed creative. Novelty can be minimal and may  not be considered sufficient; the definition is consequently sometimes extended by a third component-- \emph{surprise} \citep{boden2004creative}. With the surge of more powerful creative \ac{AI} systems, this \enquote{standard definition} has recently been updated to fence human from \enquote{artificial creativity} via the features of authenticity and intentionality \citep{runco2023updating}. This proposal can be argued to lack behind \ac{CC} research which has long recognised both features as central challenges and identified  algorithmic proposals for their realisation \citep[cf.][e.g.]{ventura2016mere,guckelsberger2017addressing,colton2018issues,mccormack2019autonomy}. 

Unfortunately, these definitions only resolve some of the ambiguity while introducing more. In particular the meaning of authenticity and intentionality remain subject of philosophical debate, and the meanings of \emph{value}, \emph{novelty}, and \emph{surprise} vary depending on who evaluates the creativity, of what, and in which context \citep{silvia2018creativity,puryear2020creativity}. As a consequence of this conceptual ambiguity, domain-, observer- and perspective-sensitivity, no consensus exists on standardised measures to assess \ac{CC} \citep{lamb2018evaluating}, despite various efforts \citep[e.g.][]{Ritchie2019}.

\subsubsection{ALife, Open-Ended Evolution, and Open-Endedness}

As a whole, the field of \ac{ALife} aims to recreate aspects of real and hypothetical living systems from scratch, i.e.~artificially. These endeavors fall under one or more of the following three categories: wet experiments (such as chemical evolution), hardware experiments (such as robotics), and software experiments (such as evolutionary models). 

It is important to note that the concept of open-ended \emph{evolution} (\acs{OEE}\acused{OEE}), arising from the ALife community, predates the generalization to \acf{OE} more broadly. Foundational descriptions of \ac{OEE} originate from many pre-millennium works. Specifically, \cite{bedau1998four} finds the phrase ``open-ended evolution'' previously used by \citet{lindgren1991evolutionary} and \citet{ray1996approach}. \citet{bedau2000open} then establishes determining ``what is inevitable in the open-ended evolution of life'' as a grand challenge for \ac{ALife}. Although several more grand challenges have been identified, this still emphasises the importance of \ac{OE} and \ac{OEE} to \ac{ALife} research. Notably different to the concept of creativity in \ac{CC} though, \ac{OE}/\ac{OEE} are better understood as supporting ALife's central goal of studying and replicating life and life-like processes, rather than as the central goals themselves.

After nearly two decades of work on the problem, \citet{packard2019openended} note that evolution's dynamic production of diversity ``has led life's creative productivity to be called Open-Ended Evolution (OEE) in the field''\footnote{The curious reader is directed towards \cite{packard2019openended} for an excellent and more extensive introduction to \acf{OEE} and \acf{OE}.}. 
The typical approach towards understanding and synthesizing the phenomenon has been to create virtual worlds imbued with features of the natural world thought to be key to unbounded generation (of whatever things, be it code snippets or embodied agents), and then performing systemic experiments probing exactly how long the generative process lasts and for what reasons it might fail. Notable systems establishing the foundation for current research include Tierra \citep{ray1996approach}, Avida \citep{ofria:alife04}, PolyWorld \citep{yaeger:alife94}, Geb \citep{channon:ijss00}, Cosmos \citep{taylor:thesis}, Division Blocks \citep{spector:gecco07}, Evosphere \citep{miconi:cec05}, and Chromaria  \citep{soros:alife14}. It should be noted that this list is not exhaustive.

Throughout these decades, many definitions of \ac{OEE} have been proposed. To only name a few, it has been defined as:
\begin{itemize}
\item a system in which the ``number of possible types by far exceeds the number of individuals (copies, sequences, etc.) in a plausible (realistic) population''\\\citep{msmith:book95},

\item the ``on-going and indefinitely creative production of significantly new kinds of adaptive responses to significantly new kinds of adaptive challenges and opportunities''\\ \citep{bedau1998four},
\item ``a system in which components continue to evolve new forms continuously, rather than grinding to a halt when some sort of ‘optimal’ or stable position is reached''\\\citep{taylor:thesis}
\item ``a process in which there is the possibility for an indefinite increase in complexity'' \citep{ruizmirazo:bp08}
\end{itemize}

\citet{dolson2015holdingback} turn the question of what \ac{OEE} \emph{is} (and, accordingly, how it should be measured) on its head and ask instead what \ac{OEE} \emph{is not}. Key indicators include a lack of change in the population, a lack of novelty, a lack of increasing complexity, a lack of ecosystem diversity, and a lack of major transitions. This pluralistic account then admits more differentiation, allowing us to consider a wide variety of systems ranging from definitely-not-open-ended, lacking all of these features, to very-strongly-open-ended. However, the steps in this gradient (and the gradient in general) remain unclear and difficult to categorize at present.

The remainder of this section shifts from discussing OEE specifically to discussing OE in the broader sense. As noted earlier in this section, OE can be considered a property of systems that might be evolutionary or not. A few notable works functionally decouple OE and evolution in their definitions. \citet{lehman2008exploiting} establish ``a link between the challenge of solving highly ambitious problems in machine learning and the goal of reproducing the dynamics of open-ended evolution in artificial life''. They refer to ``open-endedness in natural evolution'', hinting at the possibility that other kinds of open-endedness might exist apart from the biological or Darwinian kind. At the same time, \cite{miconi2008road}, writing about coevolution, refers to the phenomenon as ``the simplest way to implement a more natural, possibly open-ended form of evolution''. What is significant about this framing is the implicit statement that not all kinds of evolution are necessarily open-ended. \citet{pattee2019evolved} additionally note that ``open-endedness is often considered a prerequisite property of the whole evolutionary system and its dynamical behaviors'', further explicitly separating the concepts of \ac{OE} and evolution. 

Much like creativity, the concept of \ac{OE} in \ac{ALife} is ambiguous and, arguably even more so than creativity in \ac{CC}, its definition is subject to ongoing debate. \cite{banzhafDefiningSimulatingOpenended2016} summarize historical disagreements on what constitutes \ac{OE}, noting that definitions tend to focus on the ``continual (unbounded) creation'' of either novelty or complexity.
Also similar to creativity in \ac{CC} though, these definitional components (e.g. novelty) are in themselves ambiguous and nearly impossible to untangle from context. As another similarity, there is a distinct lack of consensus for ``gold-standard'' quantitative measures for open-endedness \citep{taylor2016open}.~This makes it difficult to identify and consequently analyze the  behaviors/dynamics of new, innovative ALife models that agree with a field-wide consensus. \cite{song2022little} notes the ``disjunction [in ALife] between the conceptual definitions, operational definitions, and the actual implementation of measurement'' of OE. As such, it remains an open challenge to (1) develop a theory for open-endedness that easily translates to measures and algorithmic design in silico and (2) imitate and recreate real open-ended systems like biological evolution artificially. 

Some of the ambiguity in conceptualization and measurement may not be symmetrical between \ac{ALife} and \ac{CC}, offering an opportunity to mitigate these challenges in \ac{ALife}, and similarly address open questions in \ac{CC}. To exploit these opportunities though, we must better understand the connection between creativity and \ac{OE}/\ac{OEE}. This need is also demonstrated by the undifferentiated mentioning of creativity in writing about \citep[e.g.][]{packard2019openended} and definitions of \ac{OE}/\ac{OEE} \citep[e.g.][]{bedau1998four}, also quoted in this section.

\section{Early Connections: OE \& Creativity}
\label{sec:related_work}

As our first contribution, we provide an up-to-date review of connections between \ac{OE} and creativity previously drawn in related work, in chronological order. First, though, we clarify the review's scope and which work has been excluded. 

\subsection{Excluded Literature}

Explicit discussions on the relationship of creativity and \ac{OE} are scarce and we exclude any work that merely draws implicit connections and does not explicate the involved concepts from this review. For instance, \citet{stanley2017open} do not directly compare creativity and \ac{OE}, but on two occasions amalgamate the concepts into ``open-ended creativity''. \ac{OE} is framed as a modality of a process (creativity) that could just as well be bounded, akin to traditional problem-solving. Similarly, \citet{lehman2020creativityevolution} assert that evolution is inherently creative, but lacks the foresight and intentionality of human creativity. Here, \ac{OE} is not being directly related to creativity, and human creativity is not further defined. \citet{lehman2020creativityevolution} and \citet{wiggins2015evolutionary} draw connections between creativity and evolution, but neither of these investigations connects creativity with \ac{OE} as a property of evolutionary processes. As a final example, \citet{boden2015creativity}, who has had a major influence on the development of \ac{CC} as a research endeavor, finds parallels between three types of creative processes (combinational, exploratory, and transformational -- defined later) and the research matter of \ac{ALife}, but does not relate these to \ac{OE} specifically. As discussed below, this was later mended by \citet{taylor2019evolutionary, taylorEvolutionaryInnovationViewed2021}.

We also exclude work that focuses on \ac{OE} as a property of tasks/exercises/problems -- which have more than one right answer, solution or outcome and can be completed in different ways \citep[e.g.][]{epstein2011exercises}. Open-ended \emph{tasks} are often cited as a necessary condition for the emergence of creativity. Closely related,  open-ended \emph{environments} \citep[e.g.][]{guckelsberger2016supportive} can be considered as the frames within which such tasks can be established and solved. We exclude both types of related work as they are not compatible with our focus on \ac{OE} as a \emph{process} (see below); we consider a \emph{task} akin to an objective which might be defined \emph{a priori} by an observer external to the system such as a person aiming to assess the other's creativity. 
We argue that connections between creativity and \ac{OE} as a subject of \ac{CC} and \ac{ALife} research can best be drawn by focusing on the fields' effort to synthesize a creative or open-ended \emph{process} (see Background). This process binds agent embodiment and environment together in one system and is potentially constrained by a task or giving rise to tasks (cf.~creativity as problem-finding). In this view, agent embodiment, environment, and task represent variables to e.g. benchmark the generality of this synthesized process against. We leverage literature on creativity and \ac{OE} as a process to anchor our investigation and discuss systems perspectives on creativity and \ac{OE} as a conceptual commonality below. As a consequence, we notably exclude literature from educational studies and creativity research investigating the potentiality of different tasks and environments to assess and exercise human creativity. 

\subsection{A (Very) Short History of Early Connections}

To the best of our knowledge, \citeauthor{taylor2002creativity} was first to draw connections between creativity and \ac{OE} already in \citeyear{taylor2002creativity}. Following the observation that \enquote{most [previous evolutionary ALife systems] are only capable of producing innovations of the \enquote{more-of-the-same’ variety}}, he distinguishes such \enquote{basic} from \enquote{creative} \ac{OEE}. Here, creativity is used as a qualifier of open-ended processes in evolution capable of \enquote{fundamental novelty} such as the \enquote{ability of individuals to interact with their (biotic and abiotic) environment with few restrictions, and to evolve mechanisms for sensing new aspects of this environment and for interacting with it in new ways}. Taylor highlights that creativity is distinct from \ac{OE}, and that \enquote{a system capable of open-ended evolution is not necessarily creative}. He then elaborates on system design decisions shaping the potential for such fundamental novelty to emerge, comprising the design of the environment, interactions between individuals, and the individuals themselves. By doing so, he implicitly draws another connection to creativity, in that systems theories of creativity rely on the same elements \citep{taylor2002creativity}. Crucially, Taylor uses creativity to emphasise fundamental novelty but does not explicitly discuss the value component of creativity which is considered essential in most definitions \citep{runco2012standard}.

More than a decade later, 
\citet{boden2015creativity} established further conceptual connections between creativity and \ac{ALife} more generally. We include this contribution as it influences the later discussion on connections to OE/OEE specifically. First, Boden distinguishes ``psychological creativity'' (involving human thought and activity) is contrasted with ``biological creativity'', which is described as \enquote{the ability to generate new cells, organs, organisms, or species'' and includes ``the (phylogenetic) creativity of evolution, the (ontogenetic) creativity of morphogenesis, and the (autopoietic) creation of individual cells}. She then re-applies one of her most well known distinctions of creative mechanisms \citep{boden2004creative}: \enquote{the different sorts of surprise that we experience on encountering a creative idea correspond to different mechanisms for producing novelty. Those mechanisms mark three kinds of creativity: combinational, exploratory, and transformational}~\citep{boden2015creativity}. More specifically: 
\begin{itemize}
\item \textbf{Combinational creativity}: the creation of new concepts or artifacts by combining features of existing ones. 
\item \textbf{Exploratory creativity}: traversing a space of such concepts or artifacts (labelled \enquote{conceptual space}) to construct or discover new and valued instances. 
\item \textbf{Transformational creativity}: reaching new concepts or artifacts that only become accessible by altering the rules defining the space itself \enquote{so that thoughts are now possible which previously (...) were literally inconceivable}\\\citep[][p.~6]{boden2004creative}. 
\end{itemize}
\citet{boden2015creativity} emphasizes that these \enquote{are analytical distinctions between different psychological/biological processes. They are not intended as overall descriptions of the resulting forms} (or product, or concept, or artifact) resulting from a process, as a particular form may have been created by multiple types of processes. Many process theorists consider transformational creativity as the most profound \citep[cf.][]{lamb2018evaluating}, including Boden herself, who notes that only transformational creativity allows for the most radical forms of surprise. She considers surprise to be a defining feature of creativity, next to novelty and value \citep{boden2004creative}.

Applied to \ac{ALife}, combinational creative mechanisms produce new combinations of existing forms, such as new kinds of macromolecules (sugars, proteins, etc). Exploratory creative mechanisms produce variants of particular structure \textit{types} that have already been expressed, such as a variations of a type of protein. In other words, it identifies categories or classes of objects and explores variants in that particular category. Finally, transformational creative mechanisms produce forms that are impossible under a pre-supposed possibility space but are physically realized under a new transformational possibility space. Especially the last type is of interest to \ac{ALife} researchers because it is closely related to discussions about \ac{OE} processes that continually transform the adjacent possible space of forms that are physically realizable \citep{kauffman2019world}.

While Boden's  \citeyearpar{boden2015creativity} account considers \ac{ALife} more generally, \citet{sorosevolution2016} soon thereafter drew more specific connections to \ac{OEE} (but not \ac{OE}). They highlight a potential connection between \citeauthor{taylor2016open}'s \citeyearpar{taylor2016open} primary categories of hallmarks of \ac{OEE} (ongoing adaptive novelty, and ongoing growth of complexity) and Boden's \citeyearpar{boden2004creative} concepts of exploratory and transformational creativity. Additionally, they raise three open questions about the relationship between \ac{OEE} and creativity theory: (1) (mentioned earlier) Are there nontrivial differences between creativity and \ac{OEE} or are they fundamentally the same concept? (2) Is there any utility in formally pursuing the nature of the relationship between evolution and creativity? (3) Do there exist metrics and practices for studying computational creativity (or creativity in general) that could help identify and quantify hallmarks of open-ended evolution? They thus appeal to the \ac{CC} and \ac{ALife} struggles to define and measure creativity and \ac{OE}, respectively.

Closely related, \citet{taylor2019evolutionary} relabels and connects different kinds of novelties defined by \citet{banzhafDefiningSimulatingOpenended2016} with Boden's types of creativity. \citeauthor{banzhafDefiningSimulatingOpenended2016}~distinguish three types of novelty based on whether they require changes to a system’s model or meta-model. The model describes the system's behavior through concepts which are themselves defined in a meta-model. Exploratory novelties (\citeauthor{banzhafDefiningSimulatingOpenended2016}: variation) result from applying the current dynamics of a system (summarised as a model) to multiple possible states. Expansive novelties (\citeauthor{banzhafDefiningSimulatingOpenended2016}: innovation) result from applying similar models that share concepts present in the current meta-model. Lastly, transformational novelties (\citeauthor{banzhafDefiningSimulatingOpenended2016}: emergence) can only result from models made of entirely new concepts, which necessitates a change in the meta-model description of the system's dynamics. In a footnote, \citet{taylor2019evolutionary} notes that these three types ``fit closely with Boden’s concepts of exploratory, combinational and transformational creativity''. He does not elaborate on this further, except for saying that the term compositional is preferred over Boden's combinational ``to emphasise that the size of structures may increase, and that the specific arrangement and connections between components might be important''. Connecting to his pioneering \citeyear{taylor2002creativity} work, 
expansive and transformational innovations map to what Taylor referred to as \enquote{creative}, and exploratory innovations to \enquote{basic} \ac{OEE} \citep{taylor2024correspondence}. Crucially, \citet{taylor2019evolutionary} interprets the three types of novelties exclusively as flavors of ongoing adaptive novelty as a hallmark of \ac{OEE} \citep[][also mentioned above]{taylor2016open}. This marks a distinction to \citeauthor{sorosevolution2016}'s \citeyearpar{sorosevolution2016} earlier mapping which also includes ongoing adaptive complexity. This ambiguity has so far not been resolved. 

Also in a footnote, \citet{taylor2019evolutionary} draws another parallel to an earlier distinction by \citet{boden2015creativity} between I- and H-creativity. I-creativity \citep[originally P-creativity,][]{boden2004creative} captures the case when an idea, action, or object is new to an individual \citep{boden2015creativity}), while H-creativity only holds when it is historically new, to any observer. \citet{taylor2019evolutionary} notes that \citet{banzhafDefiningSimulatingOpenended2016} originally defined transformational novelty (\citeauthor{banzhafDefiningSimulatingOpenended2016}: emergence) relative to the current model and meta-model, ruling out that e.g.~a major transition, which would have to be added to the meta-model, could be considered transformational again after it initially occurred. Instead, he suggests defining novelty as ``relative to the initial model and meta-model applied to an evolutionary system at its
inception'', and connects this change to \citeauthor{boden2015creativity}'s \citeyearpar{boden2015creativity} distinction between I- and H-creativity. 

\citet{song2022little} recites some of these previous connections in less detail and highlights a \enquote{loose [conceptual] equivalence} between creativity, as defined by \citet{boden2004creative} in terms of surprise, novelty, and value (e.g. usefulness, beauty, etc., and \ac{OE} -- as defined in terms of novelty, diversity, and complexity (see Background). \citeauthor{song2022little} grounds this equivalence in \citeauthor{schmidhuber2008driven}'s (\citeyear{schmidhuber2008driven}) theory of beauty-as-compressibility. Compressibility in turn is often used to estimate complexity in \ac{OE} \citep[e.g.][]{earle2021video}. Notably, \citeauthor{schmidhuber2008driven}'s theory has received arguably little approval, and \citeauthor{song2022little}'s attempt at an equivalence misses instances of creativity that are disconnected from beauty. This leaves the exact relationship between compressibility and these three components of creativity open. Despite the contentiousness of \ac{OE} definitions, \citeauthor{song2022little} also notes that requirements for \ac{OE} are weaker than for creativity according to \citeauthor{boden2004creative}'s definition, in that \ac{OE} \enquote{requires one of novelty, diversity, and complexity, while having the contingency for the other qualities}~\citep{song2022little}.

Our goal is to go beyond statements of \enquote{loose equivalence} and propose candidates for connections which further illuminate the exact nature of this match, respecting the wide array of different conceptualisations of \ac{OE} and creativity.

\section{New Connections}

To this end, as our second contribution, we put forward a preliminary list of questions, each of which proposes a potential connection between \ac{OE} and creativity. We invite researchers to engage in in-depth exploration of these similarities, differences and potential extensions to both concepts in future work. The questions were identified by the authors, who self-identify with the \ac{ALife} (LBS, AMA, OF) and \ac{CC} (LBS, SK, CG) communities, and of which at least two (LBS, CG) have contributed research to both fields. In eight remote sessions, we collaboratively constructed an affinity diagram, capturing the connections between the two concepts. Affinity diagrams represent a qualitative data analysis method used to group data into emergent categories \citep{sharp2019interaction}. Akin to self-organized focus groups \citep{morgan1996focus}, we proposed, explored, and discussed potential connections between \ac{OE} and creativity along several dimensions, such as agency, timescale, substrate, and subjectivity. 

This process was informed by the authors' detailed knowledge of each field's research heritage. As a foundation for our investigation, we first reviewed existing surveys of \ac{OE} and (computational) creativity, as well as existing work on their connections, as discussed in the previous sections. To summarize these connections, we translate each into an open research question for future research. We reference potentially related work as a starting point where applicable:

\begin{itemize}
 
 \item Could the concept of \emph{generative creativity} \citep{bown2012generative} from \ac{CC} and the Pagan creativity tradition \citep{still2016history} provide the closest equivalence match to \ac{OE} since both conceptualisations require novelty but not value, and are inspired from natural processes?
 
 \item Could the \ac{CC} concept of \emph{adaptive creativity} \citep{bown2012generative,guckelsberger2017addressing} match with \ac{OE} formulations that comprise a notion of value as adaptive success? 

 \item Value is commonly considered an essential component of creativity \citep{runco2012standard}, but its interpretation is ambiguous and its context-/domain-dependency/agnosticity subject to ongoing debate in \ac{CC} and creativity studies \citep{loughran2018computational}; are any specific conceptualisations of value exclusive to creativity or \ac{OE}, thus denying a strict equivalence match?
 
 \item Do the goals and modelling approaches in \ac{CC} and \ac{OE} differ in their requirement for continuity? More specifically, would (computational) creativity researchers consider a system to be creative even if it quickly exhausted its potential of generating novel and valuable artifacts? Could \ac{ALife}'s understanding of \ac{OE} prompt an extension of \ac{CC}'s scope toward continuously creative system? What does the scarce work on continuous \ac{CC} \citep{cook2017vision} betray? 
 
\item Could \citeauthor{wiggins:comp_creativity}' \citeyearpar{wiggins:comp_creativity,wiggins:cf_2006,wiggins:cf_2019} and \citeauthor{ritchie:csf}'s \citeyearpar{ritchie:csf} formalisations and extensions of \citeauthor{boden2004creative}'s \citeyearpar{boden2004creative} theory of creative processes reduce ambiguity in mapping proposals to different types of novelties \citep[][highlighted earlier]{sorosevolution2016, taylor2019evolutionary} as hallmark of \ac{OEE} and consequently support the evaluation of \ac{OE}? 

\item Moreover, could their mechanistic modelling of creative processes, e.g. transformational creativity as search on a meta-level (cf.~\citet{banzhafDefiningSimulatingOpenended2016}: meta-model) inform the development of \ac{OE} algorithms, e.g. those capable of emergence \citep{banzhafDefiningSimulatingOpenended2016}? Could this be further inspired and supported by existing and new, collaborative \ac{CC} algorithm proposals to realise transformational creativity \citep[e.g.][]{grace2015specific,demke2023transformational}?

 \item Could Boden's \citeyearpar{boden2004creative} three types of creative processes and (ideally unambiguous) \ac{ALife} interpretations \citep{sorosevolution2016, taylor2019evolutionary} be used as diagnostics for \ac{OE}? Would a system be considered open-ended even if it only briefly exhibited transformational-, but then mostly produced exploratory novelties?

 \item Is \citeauthor{banzhafDefiningSimulatingOpenended2016}'s \citeyearpar[][distinguishing model and meta-models]{banzhafDefiningSimulatingOpenended2016} or Boden/Wiggins/Ritchie's \citeyearpar[][distinguishing object- and meta-level]{boden2004creative,wiggins:cf_2006, wiggins:comp_creativity, wiggins:cf_2019,ritchie:csf} theory more expressive? Given that present formalisations of the latter only permit for certain modifications on the meta-level (e.g. to the traversal function), could \citeauthor{banzhafDefiningSimulatingOpenended2016}'s \citeyearpar{banzhafDefiningSimulatingOpenended2016} theory inspire extensions of the latter, or the latter inform constraints on the former?
 
 \item Having resolved questions on conceptual (dis-)similarity, under which circumstances, if at all, can creativity/\ac{OE} be used to characterize an open-ended/creative process \citep[e.g.][]{bedau1998four,packard2019openended,hughes2024openendedness}?

 \item Is creative autonomy \citep{jennings2010developing, saunders2012towards, guckelsberger2017addressing} a necessary requirement for futher narrowing the gap between creativity and \ac{OE}?
 
 \item If creative autonomy is defined (amongt others) through independence of a programmer's instructions \citep{jennings2010developing, saunders2012towards, guckelsberger2017addressing}, then what is the reference point that a system should be independent of to be considered open-ended?

\item \ac{CC} researchers have proposed autopoiesis and adaptivity \citep{saunders2012towards, guckelsberger2017addressing} as requirements and drivers of \enquote{genuine} (not designer-imposed) creative autonomy and intentional agency. Could \ac{CC} researchers adopt synthetic equivalents of these processes from \ac{ALife} to progress on this agenda? Could \ac{OE} research inform \ac{CC} of other requirements for and drivers of creative autonomy, e.g. through the lens of continuity?

\item Can research on \ac{OE} and \ac{CC} inspire the viewpoint from which the respective phenomenon is driven and evaluated? Despite few counter-proposals \citep[e.g.][]{guckelsberger2017addressing}, most \ac{CC} systems are evaluated w.r.t.~the goals of an external observer (often injected into the system) \citep{colton2008creativity,colton2012computational}. Moreover, despite the popularity of systems theories of creativity \citep{vygotsky1930, Csikszentmihalyi2014}, creativity in most, albeit notably not all \citep[e.g.][]{saunders2015computational}, systems is driven from the inside-out, and decoupled from a system's surroundings \citep{guckelsberger2021embodiment}.

\item Related, does an equivalence match of creativity and \ac{OE} require strict commitment to a systems (and embodied) theory of creativity and rejection of isolated accounts? 

 \item What is the trade-off between \ac{OE}, \ac{CC}, and controllability in a dynamical system? How are the first two concepts related to an observer's ability to control a system's dynamics? Can the relationship of either one of the first two concepts illuminate the role of the other?
 
 \item While evolution is explicitly related to the broader notion of open-endedness, in \ac{CC} it is usually used as a metaphor for algorithm design \citep{mccormack2019creative}, and to inspire studies of creativity in evolution \citep{wiggins2015evolutionary}. Could \ac{OE} research highlight another connection between creativity and evolution to inspire work in \ac{CC}?
 
\end{itemize}
Each question could only be elaborated on briefly; we encourage researchers interested in investigating them further to contact the authors for further discussion and clarification.

\section{The Road Ahead: Bridging Communities}
\label{sec:bridging_communities}

The dual pursuits of synthesizing creativity in \ac{CC} and \ac{OE} in \ac{ALife} involve understanding the rich histories of each research agenda, as briefly summarized in this paper. Our goal has not been to move toward a complete coalescence of creativity and \ac{OE} but, through intentional dialogue about both the similarities and differences in approaching each challenge, we hope to arrive at a clearer vision of how best to further our individual aims by understanding as tightly interconnected with those of \ac{CC} and \ac{ALife} as neighbouring communities. To this end, we suggest a few questions that, if attended, would likely help push both communities toward a fuller understanding of these respective fields:

\begin{itemize}

 \item Can we translate research facets and goals from \ac{CC} to \ac{OE} research, and vice versa? For instance, an interesting exercise might be to imagine ``creativity as it might be'', as previously proposed by \citet{guckelsberger2017addressing}. 

 \item Can engagement with the other disciplines' research heritage draw our attention to missing pieces in our present, own conceptualisation of creativity and \ac{OE}?
 
 \item In light of these recent expansions of OE into more mainstream AI research, can the lens of creativity theory help contextualize OE as primarily distinct from optimization?

 \item Can the language of creativity and open-endedness be used to inspire novel research in fields outside of \ac{CC} and \ac{ALife} communities? Language about open-endedness, as noted, has been lately spreading to the broader AI/ML spheres-- at least in a limited fashion. 
\end{itemize}

We note that \ac{ALife} and \ac{CC} researchers can look back at a long history of intellectual exchange. One of the earliest examples is Bentley and Corne's 2002 collection \enquote{Creative Evolutionary Systems} \citep{bentley2002creative}, followed by a landmark Dagstuhl meeting in 2009 \citep{boden_et_al:DagSemProc.09291.1} and a proliferation of cross-disciplinary work  \citep[e.g.][]{bown2009creative, bown2012generative, mccormack2012creative, guckelsberger2017addressing, mccormack2019creative}. This history bears witness of our endeavour's inspirational and transformational potential. The very positive feedback and acceptance at this conference, one of the primary venues of \ac{ALife} research, highlight the willingness of the \ac{ALife} community to take this initiative forward; separate calls by \ac{CC} researchers to bridge communities \citep[e.g.][]{cook2018neighbouring} support that it will also fall on fertile \ac{CC} ground.

\section{Conclusions and Future Work}
The engineering of \acf{CC} and \acf{OE} have been identified as the \enquote{final frontier} and \enquote{the last grand challenge} of \ac{AI} research, respectively. We set out to (1) revisit previously highlighted and (2) collect further connections between creativity and \ac{OE} as central research topics within \ac{CC} and \ac{ALife}. We identify opportunities for both communities to learn from each other's research heritage, progress, and challenges, and consequently (3) advocate the advancement of \ac{OE} and \ac{CC} research through interdisciplinary collaborations. 

To promote accessibility also from outside of these two fields, we provided short introductions and historical context for progress in both \ac{CC} and \ac{ALife}. Complementing our review of existing with fresh connections, we further support that creativity and \ac{OE}, as conceptualised in these disciplines, are not generally identical.  The new connections drawn point at nuanced but fundamental differences, and enable their detailed exploration in future work.

The focus of this paper has been conceptual, motivated by the struggle to define creativity and \ac{OE} as one salient commonality. Also common to both endeavours is the ongoing debate about standardised measures; future work should revisit measurement proposals from both fields for joint advancement, focusing on e.g.~attempts to measure creativity and \ac{OE} objectively or subjectively, and sensitive to or agnostic w.r.t. a specific domain. Finally, we see an opportunity for future work to reflect closely on which methodology and meta-goals drive the study of creativity and \ac{OE} to broaden our toolkits and scientific perspectives.

Crucially, this future work should not be done in isolation; we \textit{most importantly} sound the call for further interdisciplinary discussions and collaboration between researchers interested in (computational) creativity and \ac{OE}. It is crucial to establish a dialogue, since doing so might provide additional context, insight, and directions moving forward. We hope that such endeavors will deepen our understanding of both research challenges and further elucidate the requirements to reproduce them in artificial systems -- including reducing ambiguity in conceptualisation and measurement. 

\section*{Acknowledgements}
We thank our anonymous reviewers for their excellent feedback and support for this paper's mission, and Tim Taylor for further historical references included in this final version. LS, AA, and OW were supported by Cross Compass, Ltd.

\bibliographystyle{ijcai_style/named}
\bibliography{bibliography}

\end{document}